# DESIGN OF ATTITUDE STABILITY SYSTEM FOR PROLATE DUAL-SPIN SATELLITE IN ITS INCLINED ELLIPTICAL ORBIT

J. Muliadi[*] , S.D. Jenie[†] and A Budiyono[‡]

*Abstract*— *In general, most of communication satellites were designed to be operated in geostationary orbit. And many of them were designed in prolate dual-spin configuration. As a prolate dual-spin vehicle, they have to be stabilized against their internal energy dissipation effect. Several countries that located in southern hemisphere, has shown interest to use communication satellite. Because of those countries' southern latitude, an idea emerged to incline the communication satellite (due to its prolate dual-spin configuration) in elliptical orbit. This work is focused on designing Attitude Stability System for prolate dual-spin satellite in the effect of perturbed field of gravity due to the inclination of its elliptical orbit. DANDE (De-spin Active Nutation Damping Electronics) provides primary stabilization method for the satellite in its orbit. Classical Control Approach is used for the iteration of DANDE parameters. The control performance is evaluated based on time response analysis.*

*Abstrak*— *Secara umum kebanyakan satelit komunikasi dirancang untuk beroperasi pada orbit geostasioner. Dan kebanyakan dari satelit tersebut dirancang pada konfigurasi dual-spin prolate. Sebagai wahana dual-spin prolate, satelit harus distabilkan dari pengaruh disipasi energi internal. Beberapa negara yang terletak pada belahan selatan bumi telah menunjukkan minatnya untuk menggunakan satelit komunikasi. Karena letak negara yang berada di latituda selatan, sebuah ide muncul untuk menginklinasi satelit pada orbit eliptik. Makalah ini secara khusus membahas perancangan sistem stabilitas sikap untuk satelit dual-spin prolate pada pengaruh medan gravitasi terganggu akibat inklinasi dari orbit elipsik wahana. DANDE (De-spin Active Nutation Damping Electronics) memberikan metoda penstabilan utama pada satelit pada orbitnya. Pendekatan kendali klasik digunakan pada iterasi parameter DANDE. Unjuk kerja kendali dievaluasi berdasarkan analisis respons waktu.*

---

[*] Department of Aeronautics and Astronautics, ITB, mie_pn@yahoo.com
[†] Department of Aeronautics and Astronautics, ITB, sdjenie@ae.itb.ac.id
[‡] Department of Aeronautics and Astronautics, ITB, the author to whom all correspondence to be addressed, agus.budiyono@ae.itb.ac.id

## 1. INTRODUCTION

For a stability criterion, most of early dual-spin vehicles were designed in an oblate configuration.

However, Kaplan [8] states that launch vehicle shroud constraints limited rotor diameters. In addition, the major axis stability condition effectively limited spinning spacecraft sizes. Since U.S. Air Force successfully by-passed this limitation by operating TACSAT, many of *prolate* spinners were launch to orbit. The first communication dual-spin satellite in *prolate* configuration is INTELSAT IV.

As a case of study, this work used Palapa B2R physical data to design feedback control parameters for the attitude stability system. Near the satellite's End of Life (EOL) time, several governments of Africans and Polynesians countries have shown interest to buy and re-use Palapa B2R. Because of those countries' location in the southern latitudes, an idea emerged to incline the satellite's orbit.

When Palapa B2R is operating in its orbit, DANDE (De-spin Active Nutation Damping Electronics (Ref. [2] pp. 62-68, [1] pp. 127-129, [4], [5])) provides primary stabilization method for the vehicle, while ANC (Active Nutation Control (Ref. [1])) supplies the back-up mode. With the use of DANDE as the stabilization mode, the current paper elaborates the tuning of the feedback control parameters using classical control approach.

## 2. REFERENCE COORDINATE SYSTEM

The reference coordinate systems used in describing the satellite are the body, stability and inertial axes. Ref [9] provides the definition and illustration of those axes in detail. Body Axes with their origin at the satellite's c.g. while **Error! Reference source not found.** showed the axes in the space, as explained in Ref [9].

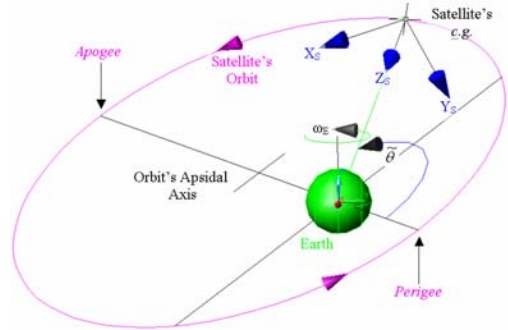

Fig. 1 Stability Reference Coordinate System

In particular, the Stability Reference Coordinate System (Stability Axes) is defined as a set of local horizon axes for the satellite. It is a target axes for the satellite's Body Axes to point its antennae to the Earth. The stability axes is presented in Fig. 1.

## 3. EULER ANGLES (ORIENTATION ANGLES)

The orientation of satellite in space with respect to a certain reference coordinate system is described by the Euler angles. Fig. describes the attitude of the satellite with respect to the Inertial Axes.

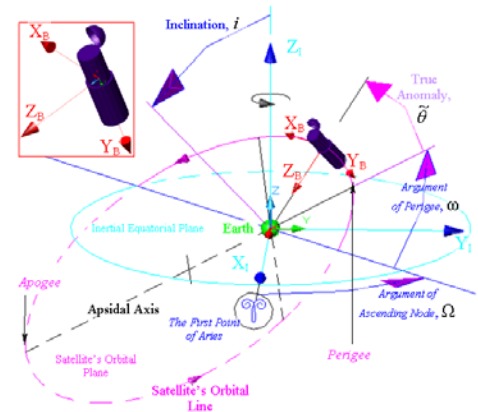

Fig. 2 Orientation of Body Axes in Inertial Axes

## 4. STATE SPACE MODEL of PROLATE DUAL SPIN SATELLITE in INCLINED ELLIPTICAL ORBIT

In order to stabilize its attitude and pointing direction, Palapa B2R uses its rotor spinning. Control moments were produced by the angular acceleration and deceleration of the rotor's spin. The satellite's motion in the yaw mode is coupled with its roll mode. In addition, the satellite's motion in the pitch mode is coupled with its yaw mode. With those couples, the satellite's attitude can be controlled by the angular acceleration and deceleration of the rotor that spun in pitch mode.

In Ref [9], the authors derive State Space Model for Dual-Spin Satellite in Stability Axes as follows,

$$\begin{Bmatrix} \dot{p} \\ \dot{q} \\ \dot{r} \\ \dot{\phi}_S \\ \dot{\theta}_S \\ \dot{\psi}_S \end{Bmatrix} = [A] \bullet \begin{Bmatrix} p \\ q \\ r \\ \phi_S \\ \theta_S \\ \psi_S \end{Bmatrix} + [B] \bullet \begin{Bmatrix} \delta e \\ \delta n \end{Bmatrix}$$

Eq. 1

where the [A] and [B] matrices are:

$$[A] = \begin{bmatrix} 0 & 0 & A_{13} & A_{14} & 0 & 0 \\ A_{21} & A_{22} & A_{23} & 0 & A_{25} & 0 \\ A_{31} & A_{32} & A_{33} & 0 & A_{35} & 0 \\ 1 & 0 & 0 & 0 & 0 & \delta n \\ 0 & 1 & 0 & 0 & 0 & 0 \\ 0 & 0 & 1 & -\delta n & 0 & 0 \end{bmatrix}$$

$$[B] = \begin{bmatrix} 0 & 0 \\ B_{21} & 0 \\ B_{31} & 0 \\ 0 & 0 \\ 0 & 1 \\ 0 & 0 \end{bmatrix}$$

By defining $\Delta_I$ as follows,

$$\Delta_I = (I_Y I_Z + I_Y I_T - I_{YZ}^2)$$

the [A] matrix elements in 1st line are:

$$A_{13} = \frac{-1}{(I_X + I_T)} \cdot (I_S \cdot \Omega_{R0})$$

$$A_{14} = \frac{-1}{(I_X + I_T)} \cdot G_X$$

the [A] matrix elements in 2nd line are:

$$A_{21} = \frac{-I_{YZ}}{\Delta_I} \cdot (I_S \cdot \Omega_{R0})$$

$$A_{22} = \frac{I_Z + I_T}{\Delta_I} \cdot \left( \frac{N \cdot K_V}{R_{d.c.}} + c \right) \cdot \left( \frac{I_Y}{I_S} + 1 \right)$$

$$A_{23} = \frac{I_Z + I_T}{\Delta_I} \cdot \left( \frac{N \cdot K_V}{R_{d.c.}} + c \right) \cdot \left( \frac{I_{YZ}}{I_S} \right)$$

$$A_{25} = \frac{-(I_Z + I_T)}{\Delta_I} \cdot G_Y + \frac{I_{YZ}}{\Delta_I} \cdot G_Z$$

the [A] matrix elements in 3rd line are:

$$A_{31} = \frac{I_Y}{\Delta_I} \cdot (I_S \cdot \Omega_{R0})$$

$$A_{32} = \frac{-I_{YZ}}{\Delta_I} \cdot \left( \frac{N \cdot K_V}{R_{d.c.}} + c \right) \cdot \left( \frac{I_Y}{I_S} + 1 \right)$$

$$A_{33} = \frac{-I_{YZ}^2}{\Delta_I} \cdot \left( \frac{N \cdot K_V}{R_{d.c.}} + c \right) \cdot \left( \frac{1}{I_S} \right)$$

$$A_{35} = \frac{I_{YZ}}{\Delta_I} \cdot G_Y - \frac{-I_Y}{\Delta_I} \cdot G_Z$$

the [B] matrix elements in 1st column are:

$$B_{21} = \frac{I_Z + I_T}{\Delta_I} \cdot \left( \frac{N}{R_{d.c.}} \right)$$

$$B_{31} = \frac{-I_{YZ}}{\Delta_I} \cdot \left( \frac{N}{R_{d.c.}} \right)$$

and

$$n = -\dot{\tilde{\theta}} = \frac{-V_\theta}{R}$$

$$\delta n = (n - n_0)$$

$$= \dot{\tilde{\theta}}_0 - \dot{\tilde{\theta}} = \frac{(V_\theta)_0}{R_0} - \frac{(V_\theta)}{R}$$

where $n$ = orbital angular velocity, shown in Fig. 1.

The more complete explanation of the use of Euler angles in describing satellite orientation is explained in Ref [9].

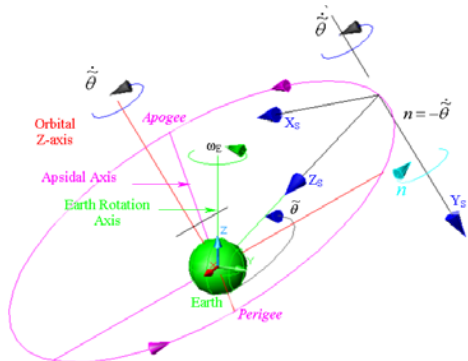

**Fig. 1 Rotation of Local Horizon Axes (Stability Axes), *n***

## 5. CONTROL STRATEGY

In elliptic and inclined orbit, the satellite attitude dynamics acts as a time-varying system. The value of $A_{14}$, $A_{25}$, $A_{35}$ and $\delta n$ will be time-varying for elliptical or inclined orbit. However, the elements of [**A**] are constant value only for circular orbit at equatorial plane. Therefore, to apply Classical Control Approach, the iteration for control parameter was held in equatorial and circular orbit.

### 5.1. Control Strategy for Longitudinal Motion

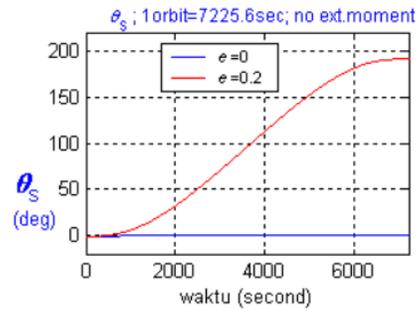

**Fig. 2 Response of $\theta_S$ without Gravity Gradient Moment Effect**

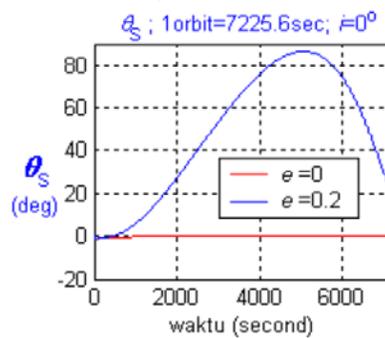

**Fig. 3 Response of $\theta_S$ in Gravity Gradient Moment Effect**

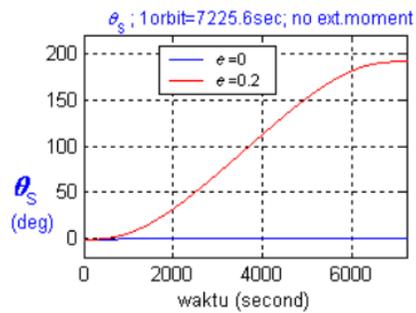

Fig. 2 is $\theta_S$ response in equatorial orbit without Gravity Gradient Moment. Fig. 3 is $\theta_S$ response in equatorial orbit in Gravity Gradient Moment Effect.

The comparison from $\theta_S$ response in elliptic orbit at

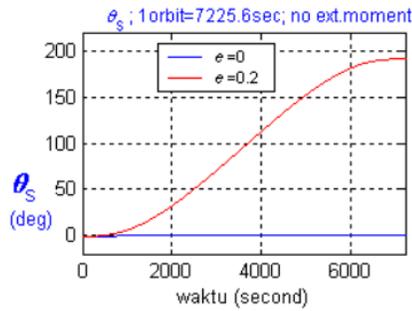

Fig. 2 and Fig. 3 showed that, "Gravity Gradient Moment tends to stabilize longitudinal motion of the prolate dual-spin satellite."

Therefore, to simplify the iteration process of longitudinal control parameter, the authors omit the Gravity Gradient Moment terms in [**A**] matrix elements. The value $A_{14}$, $A_{25}$, and $A_{35}$ are set to zero.

$$\mathbf{A} = \begin{bmatrix} 0 & 0 & 3.7113 & 0 & 0 & 0 \\ 0.49773 & -9.7138 \times 10^{-4} & -3.4402 \times 10^{-5} & 0 & 0 & 0 \\ -4.0326 & 3.3636 \times 10^{-5} & -1.1912 \times 10^{-6} & 0 & 0 & 0 \\ 1 & 0 & 0 & 0 & 0 & 0 \\ 0 & 1 & 0 & 0 & 0 & 0 \\ 0 & 0 & 1 & 0 & 0 & 0 \end{bmatrix}$$

**Eq. 2**

The value of and [**B**]'s elements are:

$$\mathbf{B} = \begin{bmatrix} 0 & 0 \\ -5.1218 \times 10^{-4} & 0 \\ 1.7735 \times 10^{-5} & 0 \\ 0 & 0 \\ 0 & 1 \\ 0 & 0 \end{bmatrix}$$

**Eq. 3**

For *q-feedback*, the damping values are too small for harmonics oscillation modes. The author uses $\theta_S$-*feedback* for longitudinal controlling shown in Fig. 4.

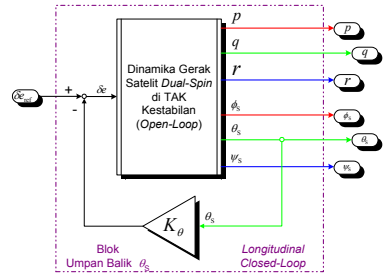

**Fig. 4** $\theta_S$-*feedback* **Diagram**

For $\theta_S$-*feedback* shown in Fig. 4, the root locus diagram is

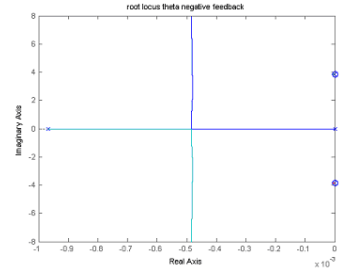

**Fig. 5** *Root-Locus* **for** $\theta_S$-*feedback* **with** $K_\theta < 0$

To increase the damping ratio for harmonics oscillation modes, the authors design a compensator. The authors start with a *pole* placement at $\frac{1}{s+1}$ (or $s_{pc}=-1$), then place a *zero* to push the harmonics oscillation *poles* more to the left.

With a *zero* placement at $s+0.2$, the locus is:

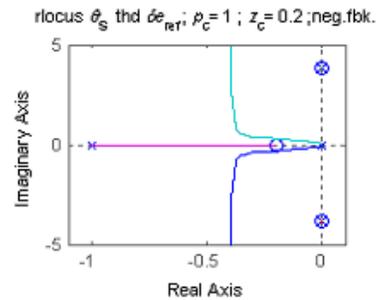

**Fig. 6** *R-Locus*, $\theta_S$, **compensated, with** $s+0.2/s+1$

With a *zero* placement at $s+0.46$, the locus is:

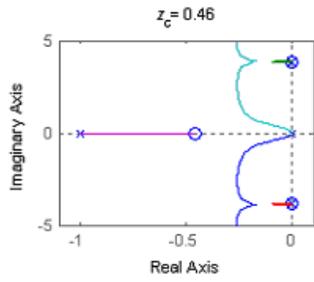

**Fig. 7** *R-Locus, θ*<sub>S</sub>**, compensated, with** s+0.46/s+1

With a *zero* placement at *s*+0.4819, the locus is:

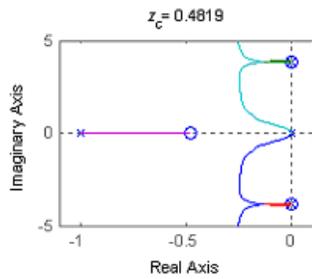

**Fig. 8** *R-Locus, θ*<sub>S</sub>**, compensated, with** *s*+0.4819/*s*+1

With a *zero* placement at *s*+0.5210, the locus is:

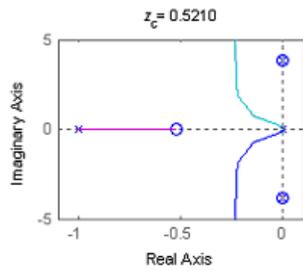

**Fig. 9** *R-Locus, θ*<sub>S</sub>**, compensated, with** *s*+0.5210/*s*+1

By detail, the compensator was chosen as follows

$$\frac{s + 0.498}{s + 1}$$

**Eq. 4**

with Root Locus Diagram (negative feedback for Fig. 4),

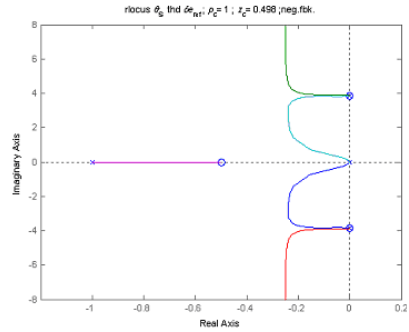

**Fig. 10** *R-Locus, θ*<sub>S</sub>**, compensated, with** *s*+0.498/*s*+1

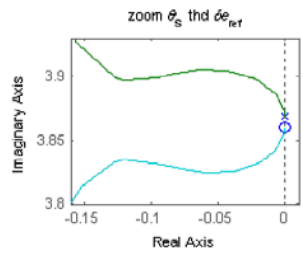

**Fig. 11 Zoom for harmonic oscillation locus (upper)**

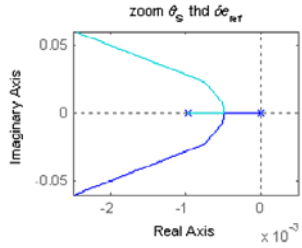

**Fig. 12 Zoom for coalescent/breakaway point**

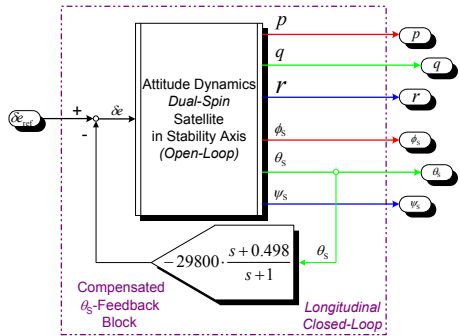

**Fig. 13 Longitudinal Controlling Diagram by $\theta_S$-feedback**

The iteration result for longitudinal controlling is $\theta_S$-*feedback*, shown in Fig. 13, with gain

$$-29800 \cdot \frac{s + 0.498}{s + 1}$$

**Eq. 5**

## 5.2. Control Strategy for Lateral Motion

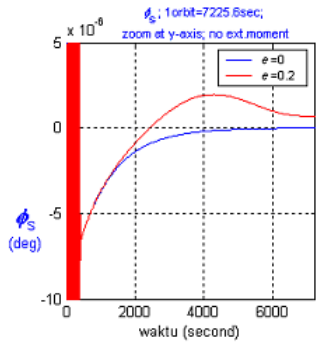

**Fig. 14 Response of $\varphi_S$ without Gravity Gradient Moment Effect**

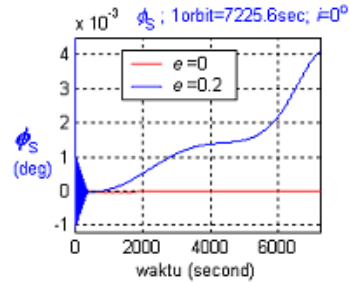

**Fig. 15 Response of $\varphi_S$ in Gravity Gradient Moment Effect**

Fig. 14 **is $\varphi_S$ response in equatorial orbit without Gravity Gradient Moment.**

Fig. 15 is $\varphi_S$ response in equatorial orbit in Gravity Gradient Moment Effect.

The comparison from $\theta_S$ response in elliptic orbit at
Fig. 14 **and**
Fig. 15 showed that, "Gravity Gradient Moment tends to de-stabilize lateral motion of the prolate dual-spin satellite."

Therefore, the iteration processes of lateral control parameter should included the Gravity Gradient Moment terms in [**A**] matrix elements. The value $A_{14}$, $A_{25}$, and $A_{35}$ are non-zero,

$$\mathbf{A} = \begin{bmatrix} 0 & 0 & 3.7113 & -6.1872 \times 10^{-7} & 0 & 0 \\ 0.49773 & -9.7138 \times 10^{-4} & -3.4402 \times 10^{-5} & 0 & 7.2937 \times 10^{-7} & 0 \\ -4.0326 & 3.3636 \times 10^{-5} & -1.1912 \times 10^{-6} & 0 & -1.3422 \times 10^{-7} & 0 \\ 1 & 0 & 0 & 0 & 0 & 0 \\ 0 & 1 & 0 & 0 & 0 & 0 \\ 0 & 0 & 1 & 0 & 0 & 0 \end{bmatrix}$$

**Eq. 6**

and the value of and [**B**]'s elements as in Eq. 3.

For *p-feedback* as lateral controlling (shown in Fig. 16),

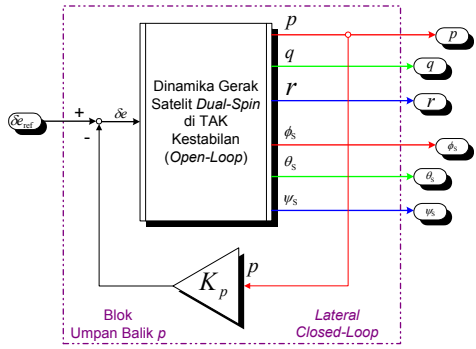

**Fig. 16** *p-feedback* **Diagram**

$$K_p \cdot \frac{(s+4.1)}{(s+25.9)\cdot(s+2.63)}$$

**Eq. 7**

and the Root Locus Diagram becomes

the Root Locus Diagram is

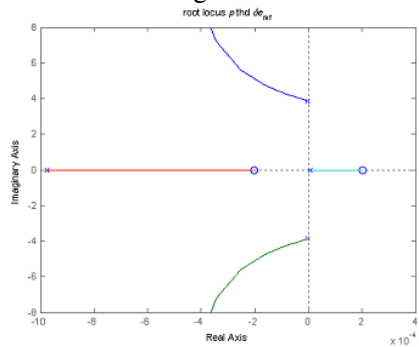

**Fig. 17** *Root-Locus* for *p-feedback* with $K_p > 0$

To push the divergence pole across imaginary axis, negative *p-feedback* applied results the Root Locus Diagram:

Pushing the divergence pole to the left side imaginary axis will destabilize the harmonic oscillation modes. Once again, the author design a compensator for *p-feedback*.

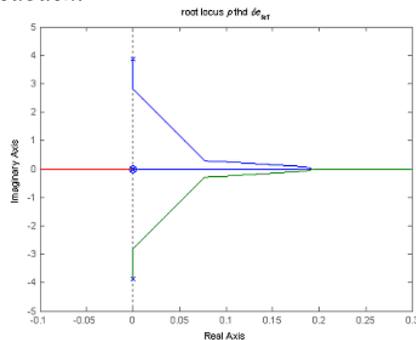

**Fig. 18** *Root-Locus* for *p-feedback* with $K_p < 0$

By trial 'n error with `sisotool` in MATLAB[®], the compensation function is,

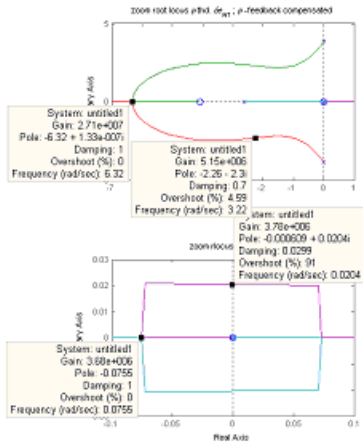

**Fig. 19** *Root-Locus* **for** *p-feedback compensated* **with** $K_p > 0$

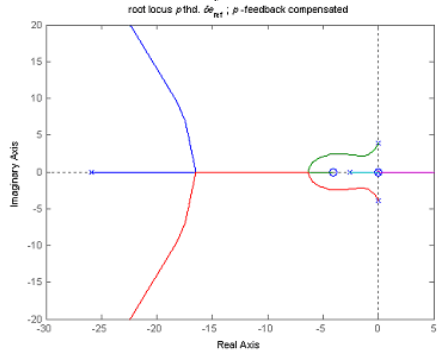

**Fig. 20** *Root-Locus* **for** *p-feedback compensated* **with several critical gains**

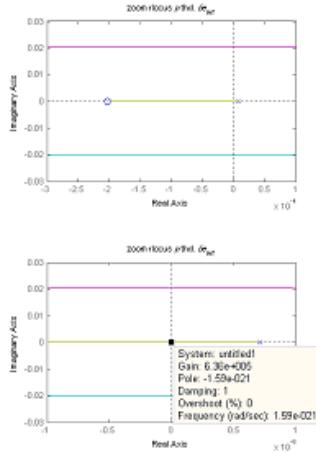

**Fig. 21** *Root-Locus* **for** *p-feedback compensated gain margin*

The iteration result for lateral controlling is *p-feedback*, shown in Fig. 22, with gain

$$1.5 \times 10^6 \cdot \frac{(s+4.1)}{(s+25.9) \cdot (s+2.63)}$$

**Eq. 8**

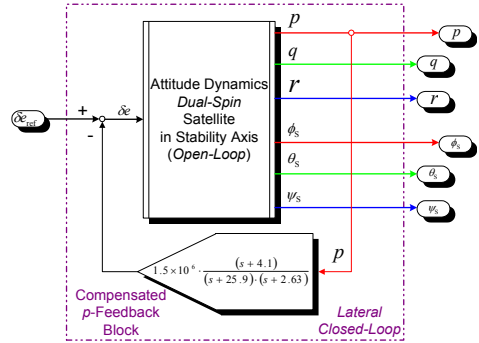

**Fig. 22 Lateral Controlling Diagram by** *p-feedback*

### 5.3. Control Strategy for Directional Motion

Because of the coupled between lateral and directional motion, the Gravity Gradient Moment Effects are also included in the iteration for Directional Control Parameters.

With [**A**] matrix elements showed in Eq. 6 and the value of and [**B**]'s elements as in Eq. 3, for the *r-feedback* as directional controlling (shown in Fig. 23),

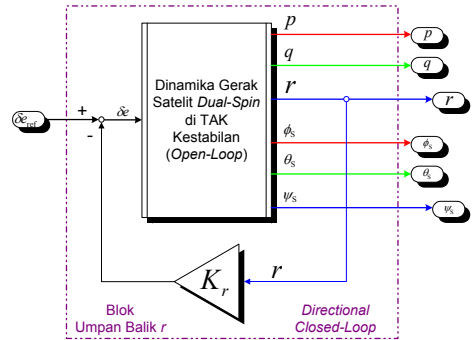

**Fig. 23** *r-feedback* **Diagram**

the Root Locus Diagram is

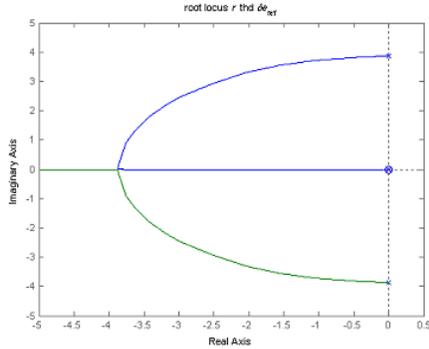

**Fig. 24** *Root-Locus* for *r-feedback* with $K_r > 0$

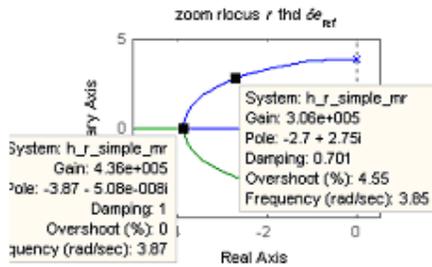

**Fig. 25** Zoom for *Root-Locus* for *r-feedback* with several critical gains

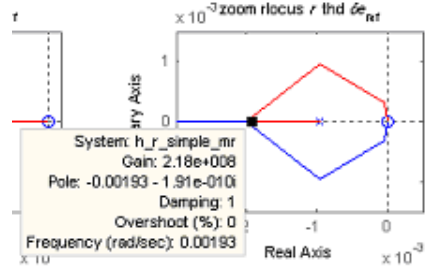

**Fig. 26** Zoom for *Root-Locus* for *r-feedback* with breakaway gain

To increase the damping value for harmonic mode, the authors use directly the Open Loop Root Locus and choose the gain as constant. The directional controlling is *r-feedback*, shown in Fig. 27, with gain

$$K_r = 300\,000$$

**Eq. 9**

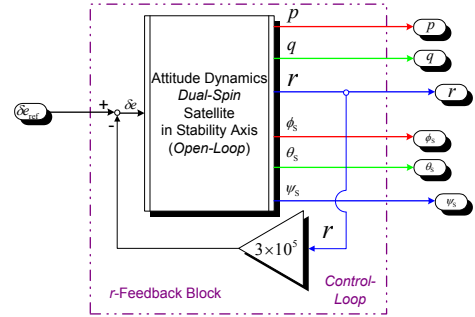

**Fig. 27** Directional Controlling Diagram by *r-feedback*

## 6. INTERPRETATIONS of CLOSED LOOP SIMULATION RESULTS

To evaluate the performance of the Attitude Control System, numerical simulations conducted with Simulink from Matlab 6.5.

The effect of the orbit's eccentricity, *e*, were induced with the variation of orbital *drift rate* in Stability Axes, *δn* together with variation of satellite's distance from the center of the Earth, *R*. Inclination of orbit will effect the simulation due to the variation of $R_{Z_I}$, the Z-Axis component of satellite's position vector in Inertial Axes. Attitude controlling was done with the deviation of armature voltage in the de-spin motor, *δe*, from its stationer value. The reference input is the command signal, $δe_{ref}$.

This paper only reports simulation results on Euler Angles since the most important thing in Satellite's Attitude Stability is to keep pointing error under pointing budget and to reduce any nutation angle.

For clarity in presenting diagrams and graphics, the concept of simulation programming and the Euler angles graphics are presented in Appendices. The concept of simulation programming diagrams describe simulation algorithm of the controlled dual-spin satellite in each mode.

### 6.1. Simulation in Longitudinal Controlled Mode

The results of attitude dynamics simulation (Fig. 28) show good damping characteristics when responding to the doublet input.

#### 6.1.1. Effect of Eccentricity in Inclined Orbit ($i = 30°$)

Graphical interpretation for Effect of Eccentricity in Inclined Orbit for $\theta_S$ (Fig. 29 and Fig. 30): Longitudinal controlling reduced $\theta_S$ deviation amplitude $-1.5°$ into damped oscillation with amplitude $0.01°$ in $30\,\text{s}$. The subsidence mode of $\theta_S$ deviation becomes oscillation mode with damping time $\pm 60\,\text{s}$. For 10 orbit period ($72256.7\,\text{s}$) the long periodic oscillation mode of $\theta_S$ were oscillating regularly from $-2 \times 10^{-6}$ to $+6 \times 10^{-6}$ deg.

#### 6.1.2. Effect of Inclination in Elliptic Orbit ($e = 0.2$)

Graphical interpretation for Effect of Inclination in Elliptic Orbit for $\theta_S$ (Fig. 31 and Fig. 32): Inclination increment will give random effect to $\theta_S$ in order to $10^{-12}$ deg for $500\,\text{s}$. For 10 orbit period ($72256.7\,\text{s}$), the increment of inclination has no effect on the response of $\theta_S$.

### 6.2. Simulation in Lateral Mode

The results of attitude dynamics simulation (Fig. 33) show good damping characteristics when responding the doublet input.

#### 6.2.1. Effect of Eccentricity in Inclined Orbit ($i = 30°$)

Graphical interpretation for Effect of Eccentricity in Inclined Orbit for $\varphi_S$ (Fig. 34 and Fig. 35): Lateral controlling reduced roll librations mode of $\varphi_S$ with amplitude $1 \times 10^{-3}$ deg becomes damped oscillation mode with maximum amplitude $1 \times 10^{-4}$ deg with damping time $25\,\text{s}$. The increment of eccentricity induced long period oscillation modes of $\varphi_S$ with smaller amplitude from its *open-loop* values (*open-loop*: $-4 \times 10^{-3}$ to $+6 \times 10^{-3}$ deg). This mode oscillates from $-1.5 \times 10^{-3}$ to $+1.5 \times 10^{-3}$ deg..

#### 6.2.2. Effect of Inclination in Elliptic Orbit ($e = 0.2$)

Graphical interpretation for Effect of Inclination in Elliptic Orbit for $\varphi_S$ (Fig. 36 and Fig. 37): In $500\,\text{s}$ of simulation, the inclination increment from $0°$ to $30°$ will amplify the $\varphi_S$ deviation amplitude to $1.5 \times 10^{-8}$ deg. For simulation time in order of orbital periode, the increasing of inclination from $0°$ to $30°$ will amplify the amplitudes of long period oscillation mode of $\varphi_S$ for $\pm 2 \times 10^{-8}$ deg.

### 6.3. Simulation in Directional Mode

The results of attitude dynamics simulation (Fig. 38) show good damping characteristics when responding the doublet input.

#### 6.3.1. Effect of Eccentricity in Inclined Orbit ($i = 30°$)

Graphical interpretation for Effect of Eccentricity in Inclined Orbit for $\psi_S$ (Fig. 39 and Fig. 40): Directional controlling reduced the yaw librations mode of $\psi_S$ and changing it into non-oscillatory mode. The $\psi_S$ response followed the doublet input in command signal $\delta e_{\text{ref}}$. The orbit eccentricity can be seen since $200\,\text{s}$ and deviates $\psi_S$ to $+4 \times 10^{-7}$ deg at $500\,\text{s}$. The orbit eccentricity also develops the undamped mode of long period oscillation. The long period oscillation mode of $\psi_S$ oscillates from $-5.5 \times 10^{-3}$ to $+1.6 \times 10^{-3}$ deg in order of orbital period. This deviation still under Palapa B2R pointing error, N-S: $0.047°$ and E-W: $0.047°$ Ref. [5].

### 6.3.2. Effect of Inclination in Elliptic Orbit (e = 0.2)

Graphical interpretation for Effect of Inclination in Elliptic Orbit for $\psi_S$ (Fig. 41 and Fig. 42): For 500 s simulation the increment of inclination from 0° to 30° will amplify the amplitude of $\psi_S$ deviation to $1.5 \times 10^{-10}$ deg. For simulation time in the order of orbital period the increment of inclination from 0° to 30° will effects the amplitude of long period oscillation mode of $\psi_S$ from $-0.75 \times 10^{-5}$ to $+0.5 \times 10^{-5}$ deg.

## 7. CONCLUDING REMARKS

In Closed-Loop simulation, the feedback parameter successfully improved the stability of satellites attitude from impulsive perturbation. Because of controlling, pitch librations on $\theta_S$ can be damped <60s, roll librations on $\varphi_S$ can be damped <30s, and yaw librations on $\psi_S$ can be damped <30s.

Longitudinal controlling with $\theta_S$-feedback also successfully reduces the effect of eccentricity. In elliptic orbit, the amplitude of $\theta_S$ oscillation reduced from $-20°$ to $+90°$ becomes 0° (±100%).

Lateral controlling with *p*-feedback also reduces the effect of eccentricity. In elliptic orbit, the amplitude of $\varphi_S$ oscillation reduced from $-4 \times 10^{-3}$ deg to $+6 \times 10^{-3}$ deg becomes $-1.5 \times 10^{-3}$ deg to $+1.5 \times 10^{-3}$ deg.

Directional controlling with *r*-feedback also cannot reduce the effect of eccentricity, but the deviation can be tolerated. In elliptic orbit, the amplitude of $\psi_S$ is oscillating from $-6 \times 10^{-3}$ deg to $+2 \times 10^{-3}$ deg. (Pointing error N-S: 0.047° & E-W: 0.047°, Ref. [7]).

The increment of inclination, *i*, has no effect on longitudinal controlled motion in the dual-spin satellite. However, increasing the inclination of orbital plane, *i*, can amplify the amplitude of long period oscillation modes in $\varphi_S$-controlled and $\psi_S$-controlled.

# 8. APPENDICES

## 8.1. Simulation in Longitudinal Mode

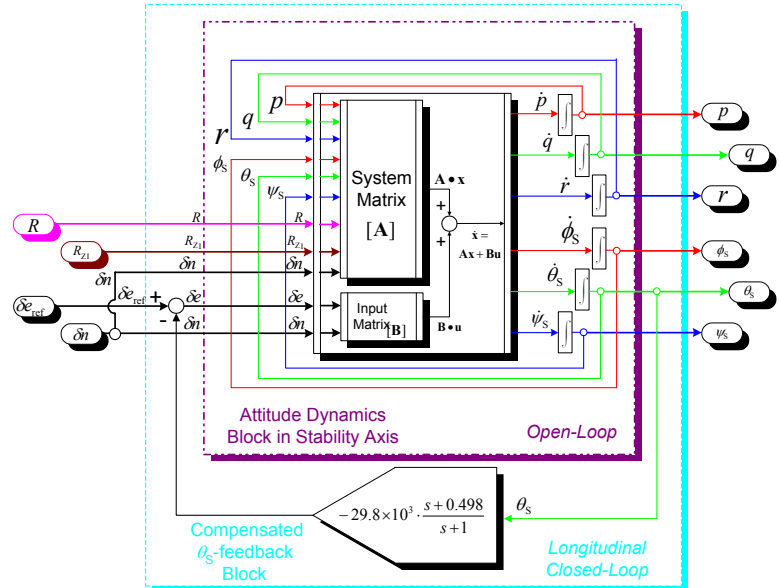

**Fig. 28 Attitude Dynamics Simulation Diagram in Longitudinal Mode**

### 8.1.1. Effect of Eccentricity in Inclined Orbit ($i$ = 30°)

◆ **Plot of $\theta_S$ because impulsive input $\delta e_{ref}$**

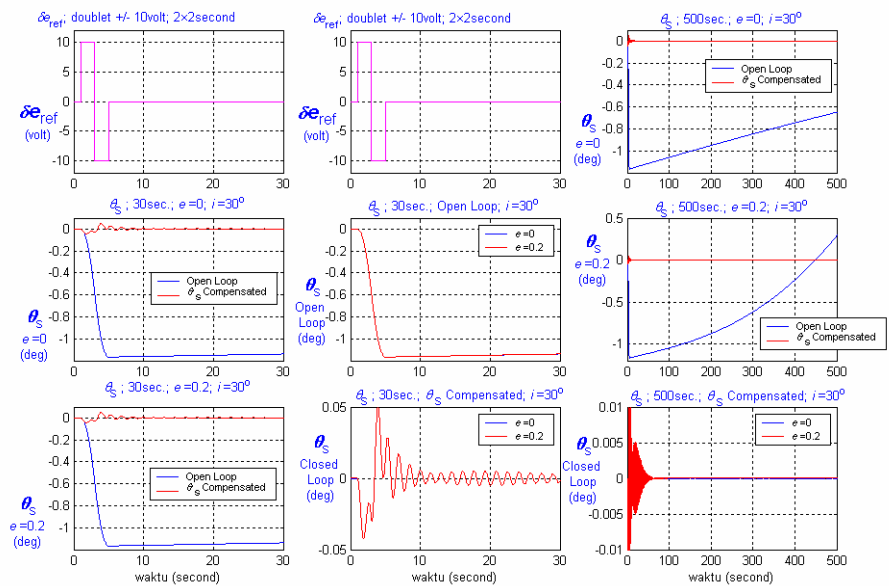

**Fig. 29 Plot of $\theta_S$; input $\delta e_{ref}$ ; $i$=30deg**

### ◆ Plot of $\theta_S$ because elliptic orbital drift input $\delta n$

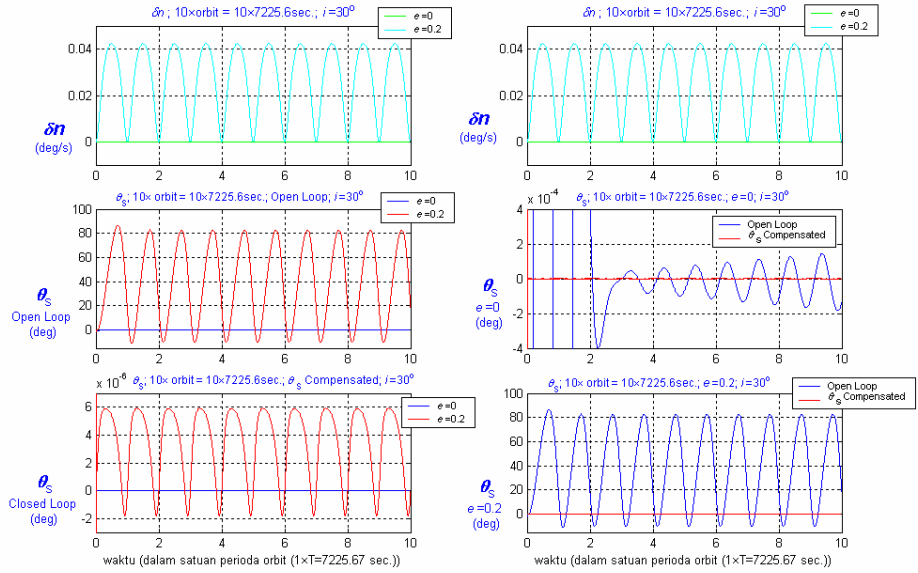

**Fig. 30** Plot of $\theta_S$; input $\delta e_{ref}$ ; $i$=30deg

## 8.1.2. Effect of Inclination in Elliptic Orbit (e = 0.2)

### ◆ Plot of $\theta_S$ because impulsive input $\delta e_{ref}$

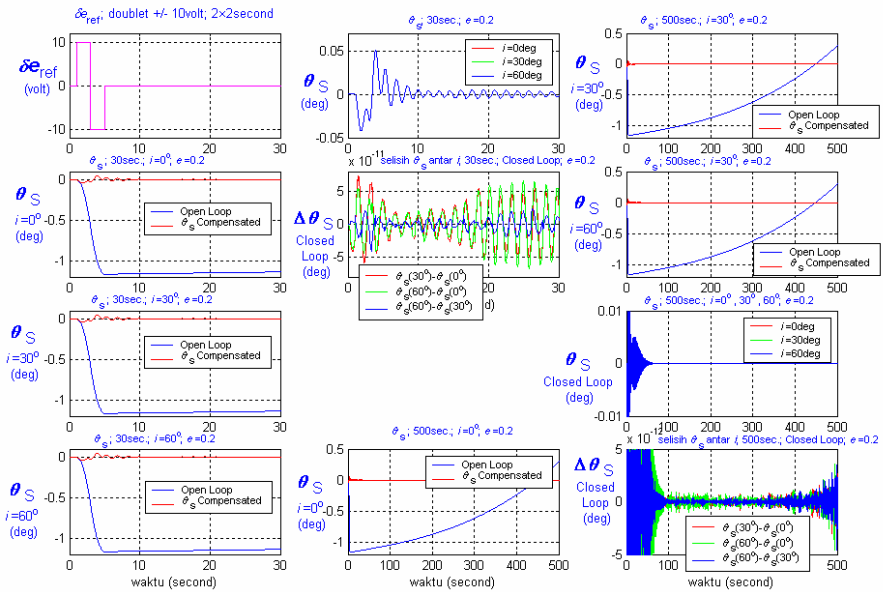

**Fig. 31** Plot of $\theta_S$; input $\delta e_{ref}$ ; $e = 0.2$

### ◆ Plot of $\theta_S$ because elliptic orbital drift input $\delta n$

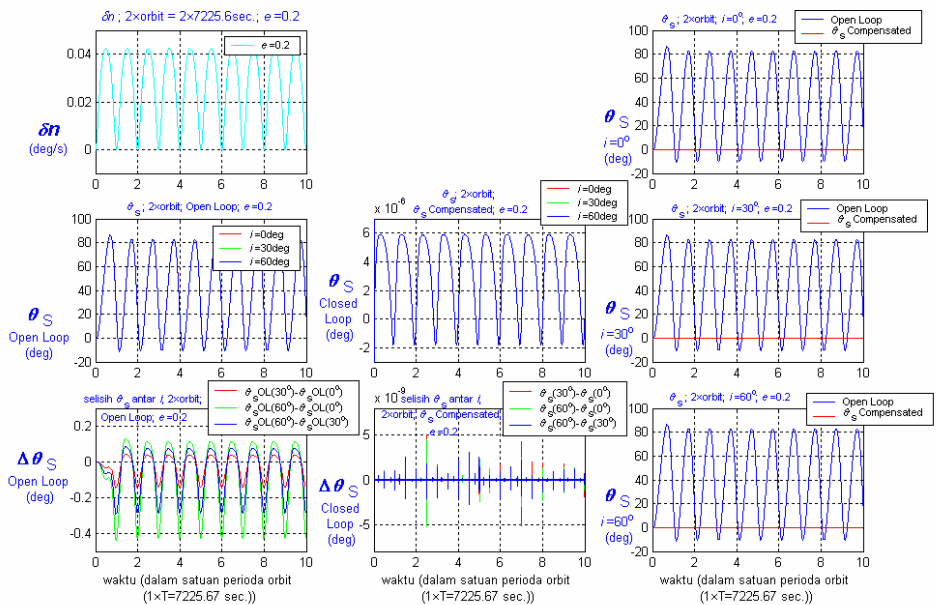

**Fig. 32** Plot of $\theta_S$; input $\delta e_{ref}$ ; $e = 0.2$; 10 Periode

## 8.2. Simulation in Lateral Mode

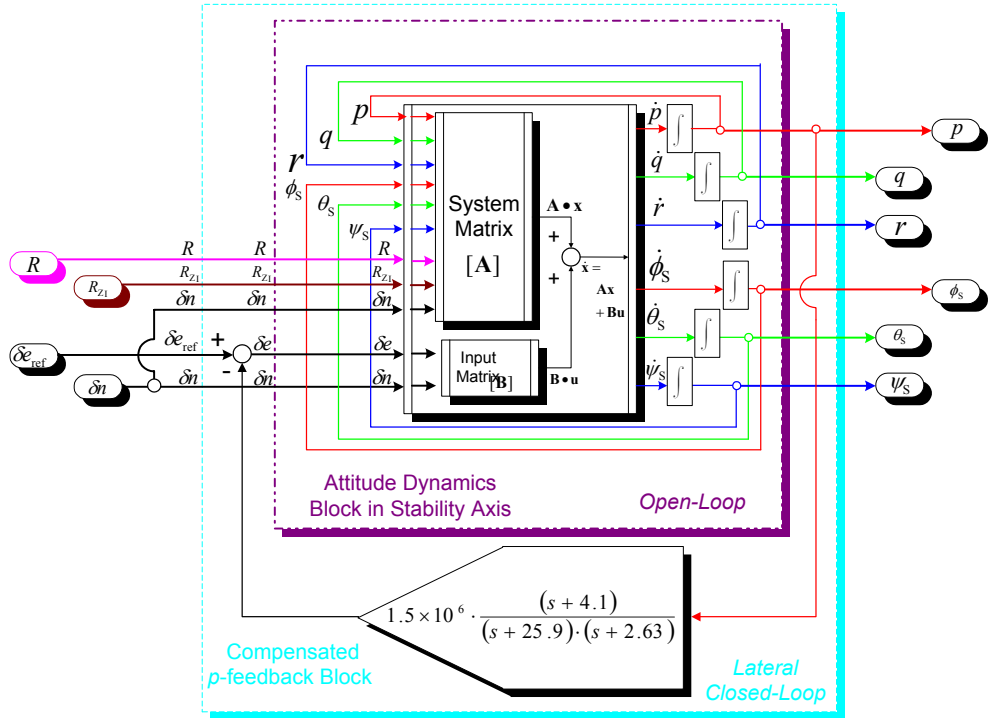

**Fig. 33 Attitude Dynamics Simulation Diagram in Lateral Mode**

### 8.2.1. Effect of Eccentricity in Inclined Orbit (*i* = 30°)

◆ **Plot of $\varphi_S$ because impulsive input $\delta e_{ref}$**

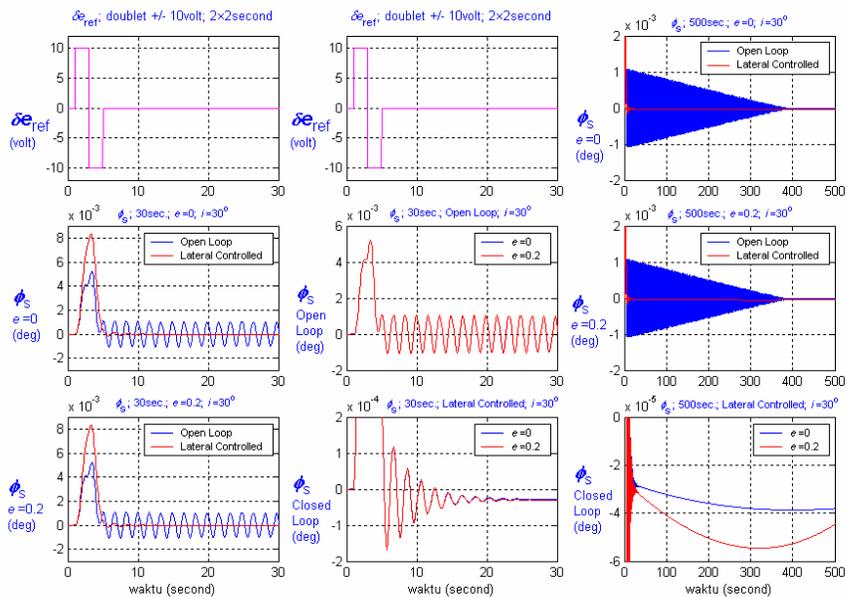

**Fig. 34 Plot of $\varphi_S$; input $\delta e_{ref}$ ; *i*=30deg**

### Plot of $\varphi_S$ because elliptic orbital drift input $\delta n$

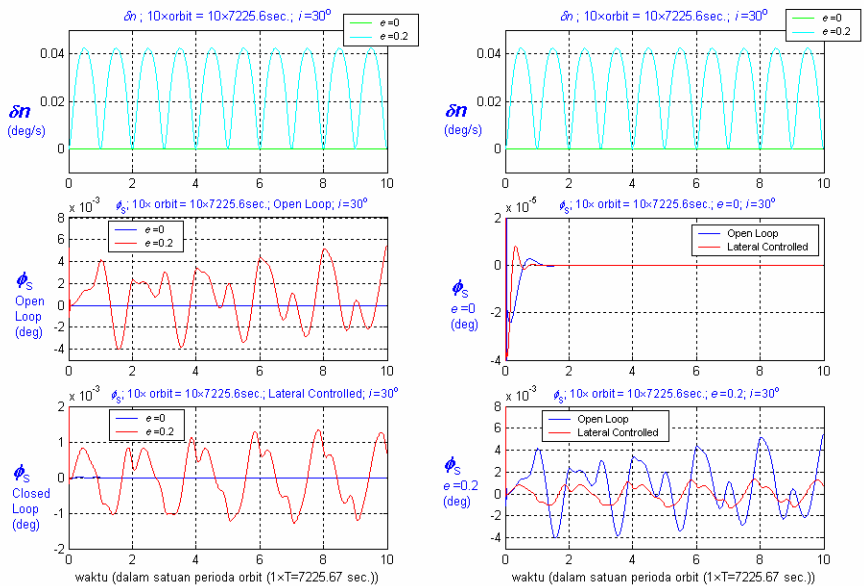

**Fig. 35** Plot of $\varphi_S$; input $\delta e_{ref}$ ; $i$=30deg

### 8.2.2. Effect of Inclination in Elliptic Orbit (e = 0.2)

### Plot of $\varphi_S$ because impulsive input $\delta e_{ref}$

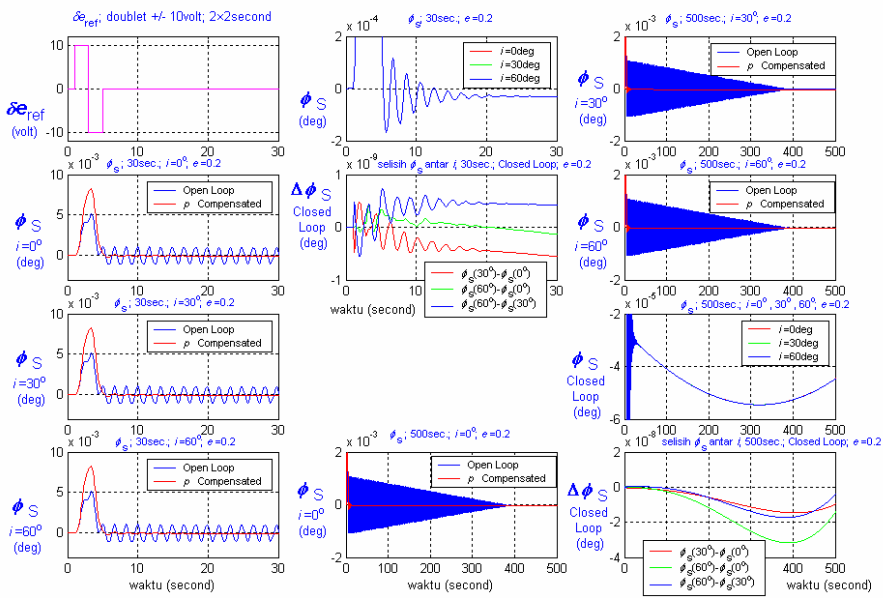

**Fig. 36** Plot of $\varphi_S$; input $\delta e_{ref}$ ; $e = 0.2$

◆ **Plot of $\varphi_S$ because elliptic orbital drift input $\delta n$**

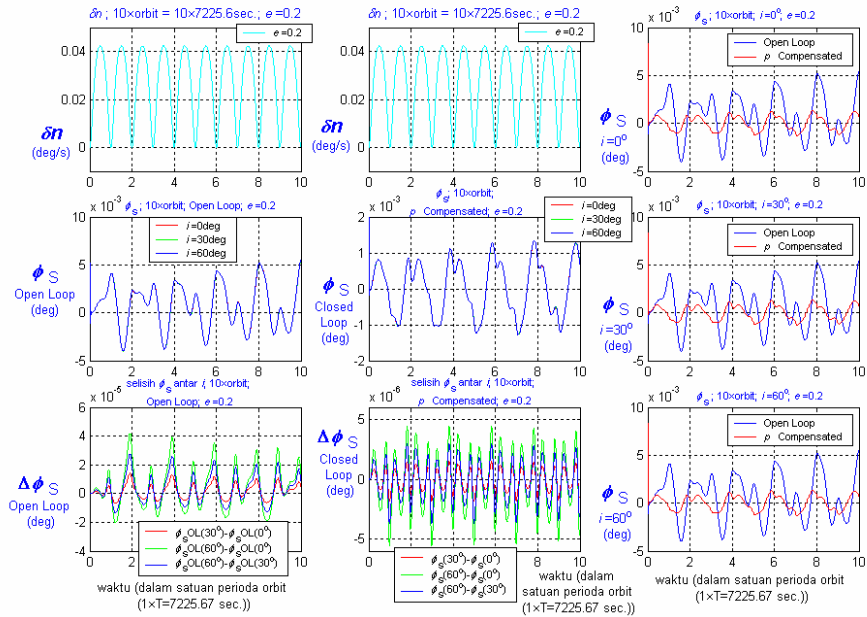

**Fig. 37 Plot of $\varphi_S$; input $\delta e_{\text{ref}}$ ; $e = 0.2$; 10 Periode**

## 8.3. Simulation in Directional Mode

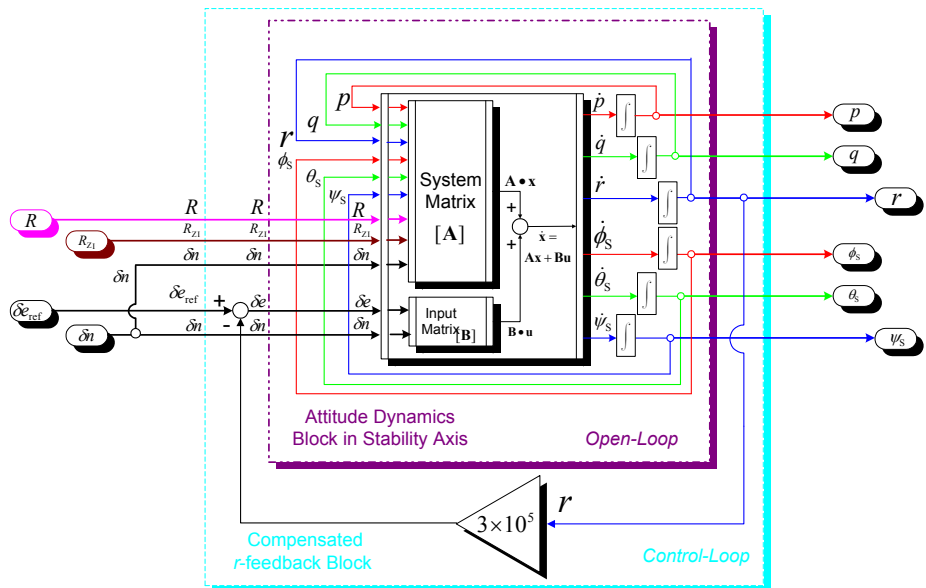

**Fig. 38 Attitude Dynamics Simulation Diagram in Directional Mode**

### 8.3.1. Effect of Eccentricity in Inclined Orbit ($i = 30°$)

◆ **Plot of $\psi_S$ because impulsive input $\delta e_{\text{ref}}$**

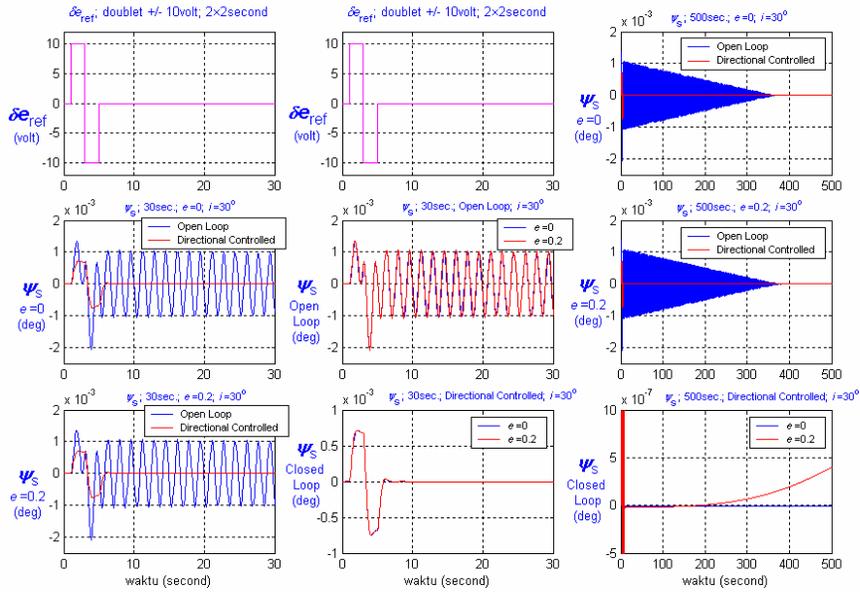

**Fig. 39 Plot of $\psi_S$; input $\delta e_{ref}$ ; $i$=30deg**

### ◆ Plot of $\psi_S$ because elliptic orbital drift input $\delta n$

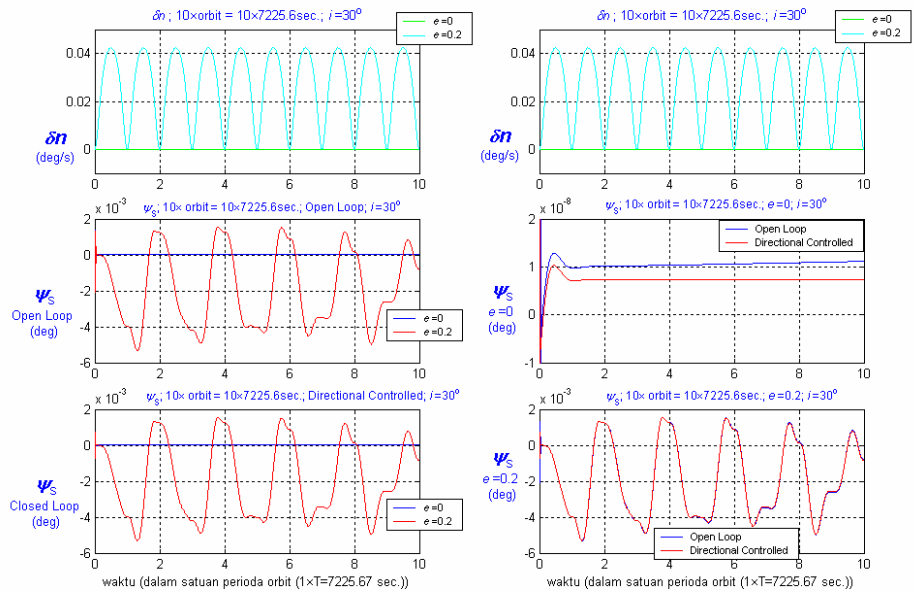

**Fig. 40 Plot of $\psi_S$; input $\delta e_{ref}$ ; $i$=30deg**

### 8.3.2. Effect of Inclination in Elliptic Orbit (*e* = 0.2)

◆ **Plot of $\psi_S$ because impulsive input $\delta e_{ref}$**

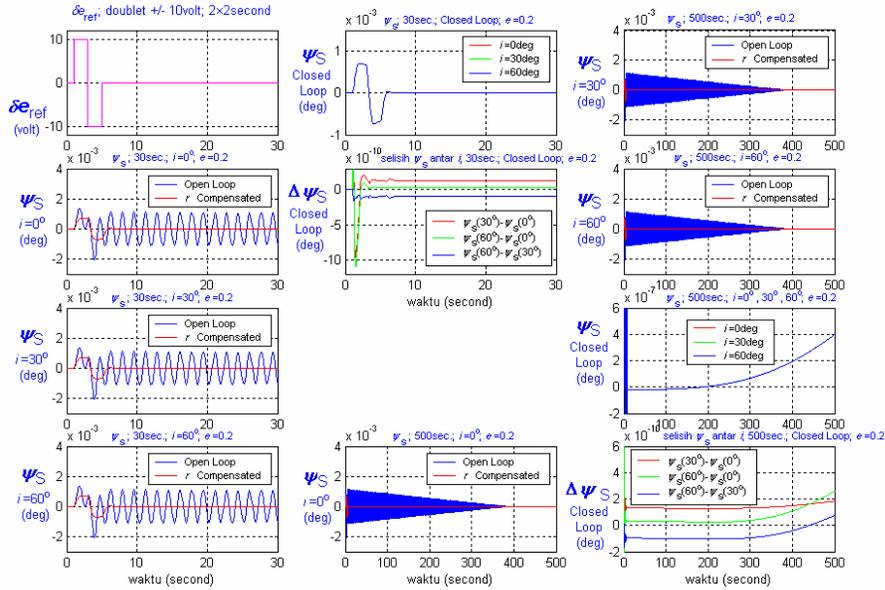

**Fig. 41** Plot of $\psi_S$; input $\delta e_{ref}$ ; *e* = 0.2

◆ **Plot of $\psi_S$ because elliptic orbital drift input $\delta n$**

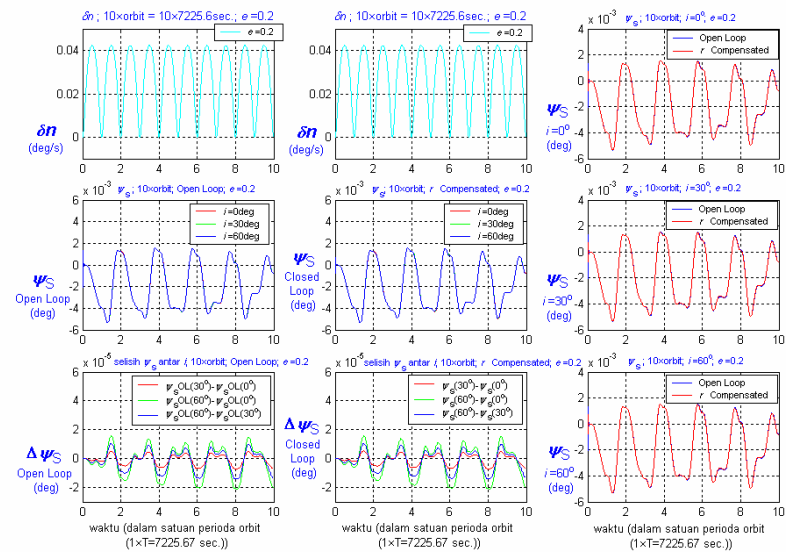

**Fig. 42** Plot of $\psi_S$; input $\delta e_{ref}$ ; *e* = 0.2; 10 Periode